\documentclass{bmvc2k}
 
\usepackage{graphicx} 
\usepackage{dsfont}
\usepackage{amsfonts}
\usepackage[linesnumbered,ruled,vlined]{algorithm2e}

\title{Grad-CL: Source Free Domain Adaptation with Gradient Guided Feature Disalignment}

\addauthor{Rini Smita Thakur}{rinithakur@iiserb.ac.in}{1}
\addauthor{Rajeev Ranjan Dwivedi}{rajeev22@iiserb.ac.in}{1}
\addauthor{Vinod K Kurmi}{vinodkk@iiserb.ac.in}{1}

\addinstitution{
 Department of Data Science and Engineering\\
Indian Institute of Science Education and Research Bhopal\\
India
}

\runninghead{Rini Thakur, Rajeev R Dwivedi, Vinod K Kurmi}{Grad-CL}

\begin{document}

\maketitle

\begin{abstract}
Accurate segmentation of the optic disc and cup is critical for the early diagnosis and management of ocular diseases such as glaucoma. However, segmentation models trained on one dataset often suffer significant performance degradation when applied to target data acquired under different imaging protocols or conditions. To address this challenge, we propose \textbf{Grad-CL}, a novel source-free domain adaptation framework that leverages a pre-trained source model and unlabeled target data to robustly adapt segmentation performance without requiring access to the original source data. Grad-CL combines a gradient-guided pseudolabel refinement module with a cosine similarity–based contrastive learning strategy. In the first stage, salient class-specific features are extracted via a gradient-based mechanism, enabling more accurate uncertainty quantification and robust prototype estimation for refining noisy pseudolabels. In the second stage, a contrastive loss based on cosine similarity is employed to explicitly enforce inter-class separability between the gradient-informed features of the optic cup and disc. Extensive experiments on challenging cross-domain fundus imaging datasets demonstrate that Grad-CL outperforms state-of-the-art unsupervised and source-free domain adaptation methods, achieving superior segmentation accuracy and improved boundary delineation. Project and code are available at \url{https://visdomlab.github.io/GCL/}.
\end{abstract}

\section{Introduction}
\label{sec:intro}
Deep convolutional neural networks have achieved remarkable success in segmenting fundus images for glaucoma diagnosis by accurately delineating the optic cup (OC) and optic disc (OD). However, these networks often experience significant performance drop when deployed on new target datasets due to cross-domain discrepancies arising from variations in acquisition protocols, scanner models, and clinical environments. Conventional unsupervised domain adaptation (UDA) methods address these discrepancies by aligning source and target data distributions using both labeled source data and unlabeled target data. In many real-world scenarios, however, the source data may be inaccessible because of copyright restrictions, confidentiality concerns, or computational limitations ~\cite{yu2023comprehensive}. This challenge is particularly acute in fundus imaging, where data from different clinical sites exhibit distinct characteristics.
Source-Free Domain Adaptation (SFDA) has emerged as a promising alternative by leveraging a pre-trained source model \(f^{s}\) (trained on \(D^{s}=(X^{s},Y^{s})\)) alongside unlabeled target images. This approach overcomes data privacy and proprietary constraints by eliminating the need for source data during adaptation. Traditional UDA methods rely on explicit alignment of feature distributions—either via specific distance metrics~\cite{m1} or through adversarial training to learn domain-invariant representations—but these strategies cannot be directly applied in the SFDA setting.\\
In SFDA, model fine-tuning is commonly achieved through self-supervised training that exploits pseudolabeling, entropy minimization, and contrastive learning. Pseudolabeling generates labels from the pre-trained source model to iteratively refine the model on the target domain. Several approaches, such as SHOT~\cite{shot}, Denoised Pseudolabeling (DPL)~\cite{dpl}, and SCNNH~\cite{tang2021nearest}, rely on computing class prototypes from target features (using techniques like weighted K-means clustering or centroid estimation) to guide pseudolabel assignment. However, these methods often suffer from noisy pseudolabels due to domain shifts and imperfect clustering. DPL is the first work which extended the SFDA model fine-tuning approach for the fundus image segmentation into two classes: optic cup and disc. The class-wise prototypes in DPL are calculated using combined features of the OC and OD for the pseudo label refinement.  We hypothesize that instead of using the integrated features from the pre-trained source model, the class-specific features of the OC and OD can be more intuitive for refinement.\\
Complementary to pseudolabeling, contrastive learning has gained prominence in SFDA by enforcing discriminative feature learning. In the SFDA setting, standard contrastive InfoNCE loss ~\cite{oord2018representation} can not be applied due to the absence of target labels. Therefore, SFDA approaches
generate positive and negative pairs in a different manner. In the absence of target labels, SFDA approaches generate positive and negative pairs using strategies such as historical source hypotheses or neighborhood information~\cite{CHEN2023110728}. The historical source hypothesis is the standard approach to handle contrastive SFDA. It contrasts the learned target embeddings of the currently adapted and historical models. HCID~\cite{huang2021model} involves the usage of memory banks for storage of outputs of historical models. UC-SFDA~\cite{CHEN2023110728} incorporates neighbourhood-guided evidence-based contrastive loss using Dempster-Shafer theory. DaC~\cite{divide} learns the global class clustering and local structures by dividing the target into source-like and target-specific samples under adaptive contrastive loss.\\
Our method (Grad-CL) includes the advantages of both pseudolabeling and contrastive learning for the fundus image segmentation into OC and OD. The inclusion of pseudo-labeling encourages discerning class-wise global assignment, whereas contrastive learning promotes inter-class separability and intra-class compactness at the local structural level. The standard thresholding technique for pseudo-labeling yields noisy labels, followed by the refining step. The uncertainty quantification and class-level prototyping are the important steps for pseudo label refinement. However, in previous works, class-level prototyping leveraged combined features obtained from the class-wise centroid estimation. Grad-CL utilizes Grad-CAM~\cite{selvaraju2017grad} to identify the features that have more impact on identifying OC and OD regions. We calculate the gradients of the segmentation regions with respect to the final convolutional feature layer of the pre-trained source model to obtain the heatmap of the optic cup and disc. Gradient-derived highlighted features of the OC and OD supply richer detail, enabling more precise centroid estimates and consequently superior pseudolabel refinement.\\
Further, we propose to utilize gradient-derived features of the optic cup and disc as positive and negative pairs for cosine similarity-based contrastive loss. The intuitive idea is to obtain the optic cup and disc features responsible for the final segmentation output and dis-align them. The dissimilarity between the optic cup and disc features helps in reducing cross-domain discrepancy. The gradient-derived cup and disc features, in collaboration with the combined features from an output of a pre-trained source model, increase the distance between disc and cup in feature space due to contrastive loss, thereby improving the adaptation performance. 
\section{Related Work}
\subsection{Unsupervised Domain Adaptation}
UDA transmits the information embedded from the  labeled source to the unseen, heterogeneous, unlabeled target domain. It mitigates the problem of performance drop due to cross-domain discrepancy and tedious pixel-level annotations. The main categorization of the UDA methods is based on~\cite{liu2022deep}: (a) alignment of the source and target distributions on the basis of specific metrics such as maximum mean discrepancy (MMD) (b) obtaining the domain-invariant features through the adversarial learning (c) image-to-image translation (d) self-training. 

In the case of fundus imaging, BEAL~\cite{beal}, pOSAL~\cite{patch}, ~\cite{feng2022unsupervised} are state-of-the-art adversarial methods for OC cup and disc segmentation. BEAL~\cite{beal}  and AdvEnt~\cite{advent} are based on the concept of entropy minimization in the adversarial adaptation network. The source images have low entropy maps, whereas the target produces high entropy maps. The idea is to enforce low entropy on the target with well-designed entropy loss or through indirect minimization with adversarial loss.  BEAL utilizes boundary predictions and entropy maps for the alignment of the source and target in an adversarial manner. poSAL~\cite{patch} is another patch-based adversarial learning method with morphology-aware segmentation loss for generating mask smoothness priors. There are hybrid methods which combine feature alignment, adversarial learning, and image synthesis. For example, the ISFA method combines feature alignment and image synthesis into a single framework for optic cup and disc segmentation~\cite{ifsa}. As part of image synthesis from intermediate latent space, the GAN generates the target-like query images. The feature consistency among source, target, and target-like queries is imposed by the content and style feature alignment module. At the final stage, output-level feature alignment generates domain-invariant features adversarially.
\subsection{Source Free Domain Adaptation}
Pseudolabeling and contrastive learning are widely incorporated in model-finetuning based SFDA methods.
The self-training by pseudo labeling implies retraining the network with pseudolabels obtained from a pre-trained source model with the target as the input. The adaptation performance depends on the quality of pseudolabels. There are different self-training based pseudo labeling methods which integrate uncertainty, confidence estimates, and evidential deep learning (EDL) to diminish the effect of poor-quality pseudo labels~\cite{kendall2017uncertainties, blundell2015weight, edl}. Pseudolabel refinement is done through uncertainty quantification through Monte-Carlo drop-out, Laplace approximation, deep ensembles, etc.

Denoised pseudolabeling (DPL)~\cite{dpl}, context-aware pseudolabel refinement (CPR)~\cite{cpr}, and PLPB~\cite{li2023robust} are important SFDA techniques catering to the pseudolabel refinement in fundus imaging. DPL generates the pseudo labels by standard thresholding, followed by pseudo label refinement with pixel-level denoising (uncertainty estimation) and class-level denoising (prototype estimation). CPR further extended DPL with the context relations refinement along with pixel-level and class-level denoising. The PLPB  method of fundus segmentation introduces additional pseudo boundary loss in addition to the pseudo label loss. The Class-Balanced Mean Teacher (CBMT)~\cite{CBMT} framework also addresses the problem of noisy pseudolabels and class imbalance with calibration loss and strong-weak augmented mean teacher model. The aleatoric uncertainty based loss and energy based loss framework refines the pseudo-labels in semi-supervised segmentation setup ~\cite{rst}. 

It has been observed that the inclusion of contrastive learning in self-training-based pseudolabeling network further improves performance. The local pixel-wise contrastive loss using pseudo labels yields good performance for the self-training-based semi-supervised segmentation~\cite{chaitanya2023local}. The robust image-classification SFDA network (PLPB) is designed using non-robust pseudo labels and pixel-wise contrastive loss for both clean and adversarial samples ~\cite{agarwal2022unsupervised}. 
\section{Proposed Method}

Grad-CL setting uses a pre-trained source model $f^{s}: X^{s} \rightarrow Y^{s}$, trained on the labelled source domain $D^{s}=(X^{s}, Y^{s})$. The labelled source data is not available during the adaptation stage. The goal of fundus imaging SFDA is to adapt the model $f^{s}$ with unlabeled target data $\lbrace {x_i ^t} \rbrace _{i=1} ^ {N_t}$, from the target domain $D^{t}$ where $x_i ^t \in  \mathbb{R}^{H \times W \times 3}$ and $ y_i^t \in \lbrace {0,1}\rbrace^{H \times W \times C} $. In this fundus image segmentation problem, the value of $C$ is two as there are two classes: OC and OD. The output adapted model $f^{s\rightarrow t}$ performs well on the target domain. The fundus image segmentation is a binary label segmentation problem into optic cup and disc regions. Grad-CL integrates two key components to achieve this:
\begin{itemize}
    \item \textbf{Gradient-guided pseudolabel refinement:} Enhances pseudolabel quality using gradient-based feature extraction and prototype estimation.
    \item \textbf{Contrastive feature disalignment:} Encourages inter-class separation between OC and OD regions by minimizing cosine similarity between class-specific gradient-derived features.
\end{itemize}
The following sections describe each component in detail.

\subsection{Gradient-guided pseudolabel refinement}

Since target data is unlabeled, we generate initial pseudolabels using the pre-trained source model. Given a target image, the model outputs per-pixel probability maps \( p_v \). We convert these probabilities into binary pseudolabels using a confidence threshold \( \gamma \): 

\begin{equation}
\widehat{y}_v^t = \mathds{1}[p_v \geq \gamma],
\label{eq:threshold}
\end{equation}

where \( \mathds{1}(\cdot) \) is the indicator function. Since raw pseudolabels are often noisy due to domain shifts, we refine them using:

\paragraph{(1) Uncertainty-Based Filtering:}
To filter out unreliable pseudolabels, we estimate pixel-wise uncertainty $(u_v)$ via Monte Carlo dropout by performing \( K \) stochastic forward passes. Given probability predictions \( \{p_v^{(k)}\}_{k=1}^{K} \), the uncertainty is computed as:

\begin{equation}
    u_v = \sqrt{\frac{1}{K} \sum_{k=1}^{K} \left( p_v^{(k)} - \bar{p}_v \right)^2},
\end{equation}
where \( \bar{p}_v \) is the mean probability at pixel \( v \). A binary mask is defined to filter out uncertain predictions:

\begin{equation}
    m_v^{u} = \mathds{1}[u_v < \eta],
\end{equation}
where \( \eta \) is the uncertainty threshold.

\paragraph{(2) Gradient-Guided Prototype:}
To further improve pseudolabel reliability, we extract class-specific gradient-derived features that highlight discriminative structures for optic cup and disc segmentation. Let \( A^k \) be the \( k \)-th feature map, and \( y^{cup} \), \( y^{disc} \) be the logits for the OC and OD, respectively. The gradient-based attention weights are computed by global average pooling over height $(h)$ and width $(w)$ as:

\begin{equation}
    \alpha_k^{cup} = \frac{1}{WH} \sum_{w}\sum_{h} \frac{\partial y^{cup}}{\partial A_{wh}^k}, \quad
    \alpha_k^{disc} = \frac{1}{WH} \sum_{w}\sum_{h} \frac{\partial y^{disc}}{\partial A_{wh}^k}.
\end{equation}

The weights $\alpha_k^{cup}$ and $\alpha_k^{disc}$ denote a partial linearization of the source model downstream from $k$ features map $A^k$, indicating the important pixels in feature maps corresponding to the OC  and OD . The gradient based coarse heatmap of OC $ e^{cup}_{GC}$ and OD $e^{disc}_{GC}$ highlighting pixels of interest is given by

\begin{equation}
    e^{cup}_{GC} = \text{ReLU}\Big(\sum_k \alpha_k^{cup} A^k\Big), \quad
    e^{disc}_{GC} = \text{ReLU}\Big(\sum_k \alpha_k^{disc} A^k\Big).
\end{equation}

The class-wise prototype calculation involves a binary object mask, binary background mask and combined features multiplied by gradient-specific features. The binary object mask $b^{ob} = \mathds{1}\left[\hat{y}^t = 1\right] \mathds{1}\left[u < \eta\right]$, binary background mask $b^{bg} = \mathds{1}\left[\hat{y}^t = 0\right] \mathds{1}\left[u < \eta\right]$, and the modified features are used for the prototype calculation. The modified features are obtained by the multiplication of gradient-derived features $e_{GCv}$ with the combined features $e_{v}$, i.e., $e_{GCv}\cdot e_{v}$. The $e_{GCv}$ denotes the gradient-derived features at pixel $v$. To refine prototypes, we first extract feature representations \( e_v \) from the model’s penultimate layer, then augment these features with their corresponding gradient-derived activations:

\begin{equation}
    e_v' = e_v \cdot e_{GCv},
\end{equation}
where \( e_v' \) denotes the enhanced feature representation incorporating gradient-based saliency.

Using the reliable pixels (filtered by uncertainty threshold \( \eta \)), we compute class prototypes:

\begin{equation}
    z^{ob} = \frac{\sum_v e_v' b_v^{ob} p_v}{\sum_v b_v^{ob} p_v}, \quad
    z^{bg} = \frac{\sum_v e_v' b_v^{bg} (1 - p_v)}{\sum_v b_v^{bg} (1 - p_v)}.
\end{equation}
The relative feature distance of object and background ($d_v^{ob}, d_v^{bg}$) is also calculated between the modified features and the class prototypes given by 
\begin{align}
d_v^{ob}=\lVert e_{v}\cdot e_{GCv} - z^{ob} \rVert_{2},  
   d_v^{bg}=\lVert e_{v}\cdot e_{GCv} - z^{bg} \rVert_{2}
\end{align}
The pseudolabel is considered noisy if the modified feature vector derived using gradient function ($\lbrace{e_v \cdot e_{GCv}}\rbrace$) is further away from the object prototype $z^{ob}$ rather than the background prototype $z^{bg}$. This distance wise denoising is merged with uncertainty based denoising to obtain the modified mask as:
\begin{equation}
m_v = \mathds{1}\left[u_v < \eta\right]\left[\widehat{y}_v^t =1\right] \left[d_{v}^{ob} < d_{v}^{bg}\right] + \mathds{1}\left[u_v < \eta\right]\left[\widehat{y}_v^t  = 0\right] \left[d_{v}^{ob} > d_{v}^{bg}\right]
\end{equation}
The pseudolabel $\widehat{y}_v^t =1$ represents the object region, whereas $\widehat{y}_v^t =0$ represents background region.
So, this binary mask  $m_v$ can be used with standard-cross entropy for the adaptation. It  leverages both uncertainty information and gradient-derived prototype estimation information.
 The final segmentation loss is computed as:

\begin{equation}
    \mathcal{L}_{seg} = \sum_v m_v \cdot \mathcal{L}_{ce,v}
\end{equation}
where \( \mathcal{L}_{ce,v} \) is the per-pixel cross-entropy loss with pseudolabels.
\begin{figure}[t]
    \centering
    \includegraphics[width=\textwidth]{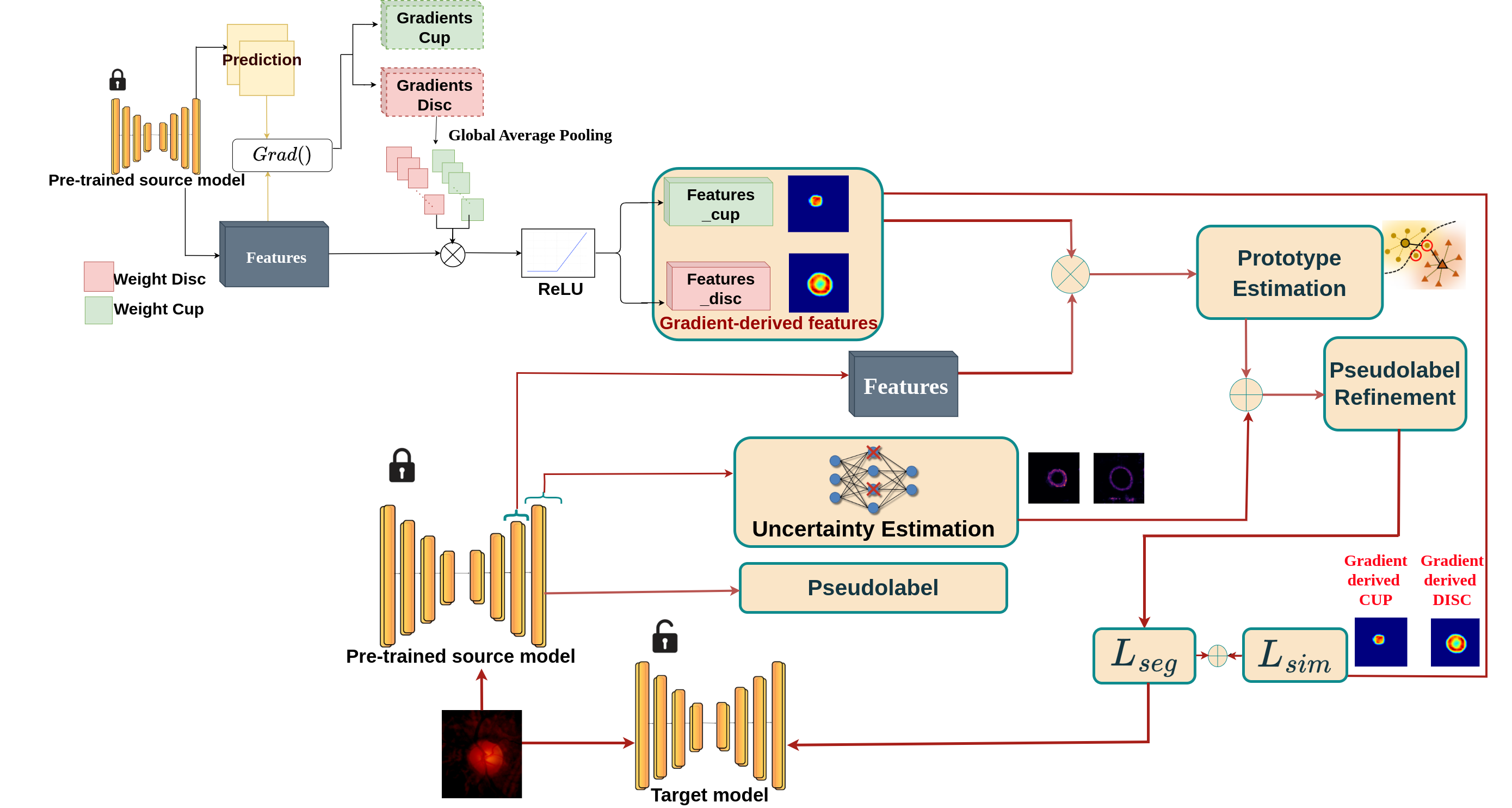}
    \caption{Overview of the proposed Grad-CL technique. The provided target image is fed into a pre-trained source model in order to generate pseudolabels through the process of thresholding. Target domain adaptation consists of two subsequent processes. a) \textit{Pseudolabel Refinement}: This process involves estimating uncertainty and deriving prototypes using gradients. b) \textit{Contrastive Loss Module}: This module is used to calculate the contrastive loss. The similarity measure $\mathcal L_{sim}$ is computed using modified gradient-derived features of the cup and disc.}
    \label{introfig}
\end{figure}
\subsection{Contrastive Feature Disalignment}
Despite refined pseudolabels, feature overlaps between the OC and OD can still lead to misclassification. To address this, we introduce contrastive learning to explicitly enforce feature disalignment. The gradient modified features highlight the key features of the optic cup and disc. We propose to utilize gradient highlighted feature, i.e., $e^{cup}_{\text{GC}}$ and $e^{disc}_{\text{GC}}$, along with the combined features obtained from the network. The disalignment of  the key features of the cup the disc by cosine similarity loss supports adaptation by creating contrast between class-specific features . Although, the combined features $e_v$ have attributes responsible for segmentation of both cup and disc. We can modify this combined features $e_v$ with the addition of gradient based feature map $e^{cup}_{\text{GCv}},e^{disc}_{\text{GCv}} $ pertaining to key regions which are actually responsible for cup and disc output. As there are regions of misinterpretation where the cup is identified as a disc and vice-versa. Therefore, the disalignment of the embeddings obtained from $\left( e_v + e^{cup}_{\text{GCv}} \right)$ and $\left( e_v + e^{disc}_{\text{GCv}}\right)$ will reduce the similarity between cup and disc features with every epoch.

We construct gradient-informed class-specific representations:

\begin{equation}
    \tilde{e}^{cup}_v = e_v + e^{cup}_{GCv}, \quad \tilde{e}^{disc}_v = e_v + e^{disc}_{GCv}.
\end{equation}

We then compute the cosine similarity loss:

\begin{equation}
    \mathcal{L}_{sim} = \frac{\tilde{e}^{cup}_v \cdot \tilde{e}^{disc}_v}{\max(\|\tilde{e}^{cup}_v\|_2 \|\tilde{e}^{disc}_v\|_2, \varepsilon)},
\end{equation}
where \( \varepsilon \) prevents division by zero.

The overall training loss is a combination of segmentation and contrastive losses:

\begin{equation}
    \mathcal{L}_{total} = \mathcal{L}_{seg} + \lambda \mathcal{L}_{sim},
\end{equation}
where \( \lambda \) balances segmentation and contrastive objectives.

Grad-CL effectively refines noisy pseudolabels using \textbf{gradient-enhanced feature extraction} while \textbf{contrastive learning} disaligns optic cup and disc representations. This results in improved segmentation performance under source-free adaptation. Figure ~\ref{introfig} depicts block diagram of Grad-CL method.

Table~\ref{tab:my-table1} summarizes the performance of Grad-CL compared to several state-of-the-art methods, illustrating the effectiveness of our refined pseudolabeling and contrastive learning strategy. The detailed steps are given in Algorithm 1.

\begin{table}[ht]
  \centering
  \footnotesize
  \scalebox{0.90}{
  \begin{tabular}{|l| l |l |l |l |l|l| l| l| l|}
    \hline
    \multicolumn{1}{|c|}{Methods} & \multicolumn{1}{c|}{S-F} & \multicolumn{2}{c|}{Optic Disc} & \multicolumn{2}{c|}{Optic Cup} & \multicolumn{2}{c|}{Optic Disc} & \multicolumn{2}{c|}{Optic Cup} \\
    \hline
      &  & Dice\%$\uparrow$ & ASD$\downarrow$ & Dice\%$\uparrow$ & ASD$\downarrow$ & Dice\%$\uparrow$ & ASD$\downarrow$ & Dice\% $\uparrow$ & ASD$\downarrow$ \\
    \hline
    \multicolumn{6}{|c|}{\textbf{RIM-ONE-r3}} & \multicolumn{4}{c|}{\textbf{Drishti-GS}} \\
    \hline
    W/o adaptation                     &       & 83.18 & 24.15 & 74.51 & 14.44 & 93.84 & 9.05  & 83.36 & 11.39 \\
    \hline
    Oracle {\cite{beal}}  &       & 96.80 & \ -    & 85.60 & \ -  & 96.40 & -     & 90.10 & -     \\
    \hline
    BEAL {\cite{beal}}    &       & 89.80 & \ -    & 81.00 & \ -    & 96.10 &       & 86.20 &       \\
    \hline
    OCDA {\cite{OCDA}}    & No    & 86.47 & 16.76 & 76.74 & 10.94 & -     & -     & -     & -     \\
    \hline
    AdvEnt {\cite{advent}}  & No    & 89.73 & 9.84  & 77.99 & 7.57  & 96.16 & 4.36  & 82.75 & 11.36 \\
    \hline
    SRDA {\cite{SRDA}}     & No    & 89.37 & 9.91  & 77.61 & 10.15 & 96.22 & 4.88  & 80.67 & 13.12 \\
    \hline
    DAE {\cite{DAE}}      & No    & 89.08 & 11.63 & 79.01 & 10.31 & 94.04 & 8.79  & 83.11 & 11.56 \\
    \hline
    TT-SFUDA {\cite{ttsfda}} & Yes   & 85.00 & 17.05 & 76.62 & 10.31 & 95.22 & 6.00  & 80.67 & 13.00 \\
    \hline
    TENT {\cite{tent}}     & Yes   & 82.92 & 23.63 & 72.95 & 14.00 & 94.06 & 7.56  & 80.12 & 13.52 \\
    \hline
    PLPB {\cite{li2023robust}} & Yes   & 92.89 & 6.52  & 77.94 & 10.07 & 96.51 & \textbf{4.01}  & 83.56 & 11.11 \\
    \hline
    DPL {\cite{dpl}}      & Yes   & 90.13 & 9.43  & 79.78 & \textbf{9.01} & 96.39 & 4.08  & 83.53 & 11.39 \\
    \hline
    \textbf{Proposed (Grad-CL)}                 & Yes   & \textbf{94.99} & \textbf{5.02}  & \textbf{80.51} & 9.50  & \textbf{96.58} & 4.09  & \textbf{84.67} & \textbf{10.28} \\
    \hline
  \end{tabular}}
  \vspace{1 em}
  \caption{Quantitative Performance of Various State-of-the-Art UDA and SFDA Methods on Target Datasets: RIM-ONE-r3 \cite{fumero2011rim} and Drishti-GS \cite{sivaswamy2015comprehensive}. Refuge \cite{ORLANDO2020101570} is taken as the Source Dataset to train the source model for adaptation.}
  \label{tab:my-table1}
\end{table}
\vspace{-2 em}

\begin{algorithm}[!h]
\smaller
\SetAlgoLined
\KwIn{
    Pre-trained source model $f^{s}(x^s \rightarrow y^s)$, \\
    Penultimate layer feature maps $A$, \\
    
    Prediction probability $p_v$,\\
    Unlabeled target data $\lbrace x_i^t \rbrace_{i=1}^{N_t} \in D^{t}$
}
\KwOut{Adapted segmentation model $f^{s\rightarrow t}(\hat{\theta}_c)$}
\For{each mini-batch of unlabeled samples $X_i^t \in D^{t}$}{
    \textbf{Self-training}\;
    $\mathcal{L}_{ce,v} \leftarrow CE(\widehat{y}_v^t, p_v)$, where $\widehat{y}_v^t$ are pseudolabels\;
    \textbf{Gradient-derived cup and disc features}\;
    Output logits $y$, $y^{cup}=y[:,0:1,:,:]$,$y^{disc}=y[:,1,:,:]$\\
    $\alpha_k^{cup} = \text{GAP}(\text{grad}(y^{cup}, A^k))$\;
    $\alpha_k^{disc} = \text{GAP}(\text{grad}(y^{disc}, A^k))$\;
    $e^{cup}_{\text{GC}} = \text{ReLU}\left(\sum_k \alpha_k^{cup} A^k \right)$\;
    $e^{disc}_{\text{GC}} = \text{ReLU}\left(\sum_k \alpha_k^{disc} A^k \right)$\;
    \textbf{Pseudolabel refinement}\;
    Define uncertainty threshold $\eta$\;
    Compute prototypes using $e^{cup}_{\text{GC}}$ and $e^{disc}_{\text{GC}}$\;
    Calculate distance vectors $d_v^{ob} = \| e_v \cdot e_{GCv} - z^{ob} \|_2$\;
    $d_v^{bg} = \| e_v \cdot e_{GCv} - z^{bg} \|_2$\;
    \textbf{Modified initial CE loss}\;
    Define modified mask $m_v$ using eq. 10\;
    $\mathcal{L}_{seg} = \sum_v m_v \cdot L_{ce, v}$\;
    \textbf{Contrastive loss}\;
    $\mathcal{L}_{sim}$ = Cosine Similarity between $\lbrace(e_v + e^{cup}_{\text{GCv}}, e_v + e^{disc}_{\text{GCv}})\rbrace$\;
    \textbf{Total loss}\;
    $\mathcal{L}_{total} = \mathcal{L}_{seg} + \mathcal{L}_{sim}$\;
}
\textbf{Output:} Trained adapted segmentation parameters $\hat{\theta}_c$\;
\caption{Source-Free Domain adaptation of fundus images using pseudolabel refinement and contrastive loss with gradient activations}
\end{algorithm}
\section{Experiments}
\textbf{Datasets:} We follow the setup in \cite{beal,dpl} by using the open-source REFUGE Challenge \cite{ORLANDO2020101570} as the source dataset and Drishti-GS \cite{sivaswamy2015comprehensive} and RIM-ONE-r3 \cite{fumero2011rim} as target datasets. The source comprises 400 annotated images (combining train and test splits), while the targets use 99/60 and 50/51 train/test splits for Drishti-GS and RIM-ONE-r3, respectively. Each image is preprocessed by extracting a \(512 \times 512\) region-of-interest that captures the cup and disc. We evaluate segmentation performance using the Dice Coefficient and Average Surface Distance (ASD).

\begin{figure*}[t]  
\centering
\begin{tabular}{ccccc}
\bmvaHangBox{\fbox{\includegraphics[width=0.15\linewidth]{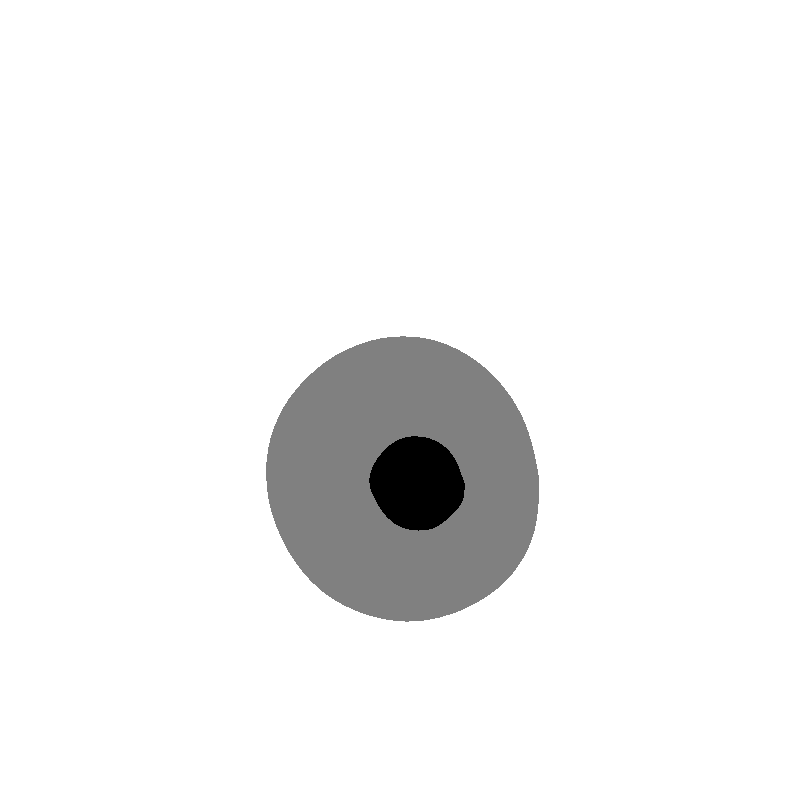}}}&
\bmvaHangBox{\fbox{\includegraphics[width=0.15\linewidth]{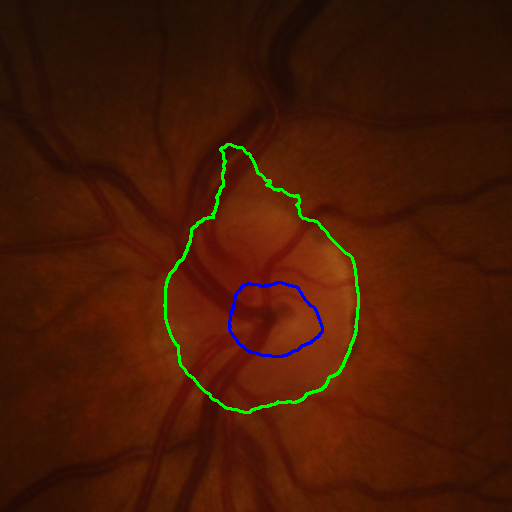}}}&
\bmvaHangBox{\fbox{\includegraphics[width=0.15\linewidth]{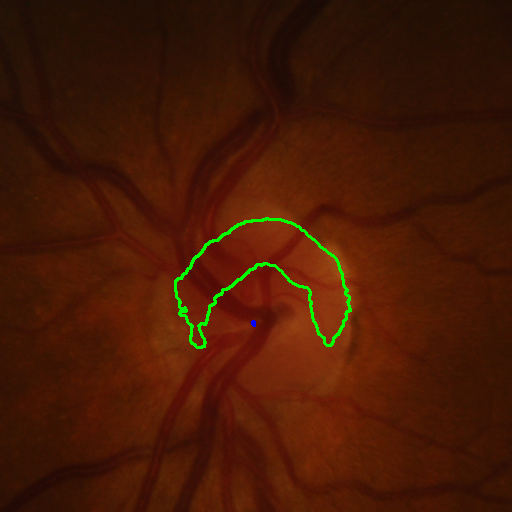}}}&
\bmvaHangBox{\fbox{\includegraphics[width=0.15\linewidth]{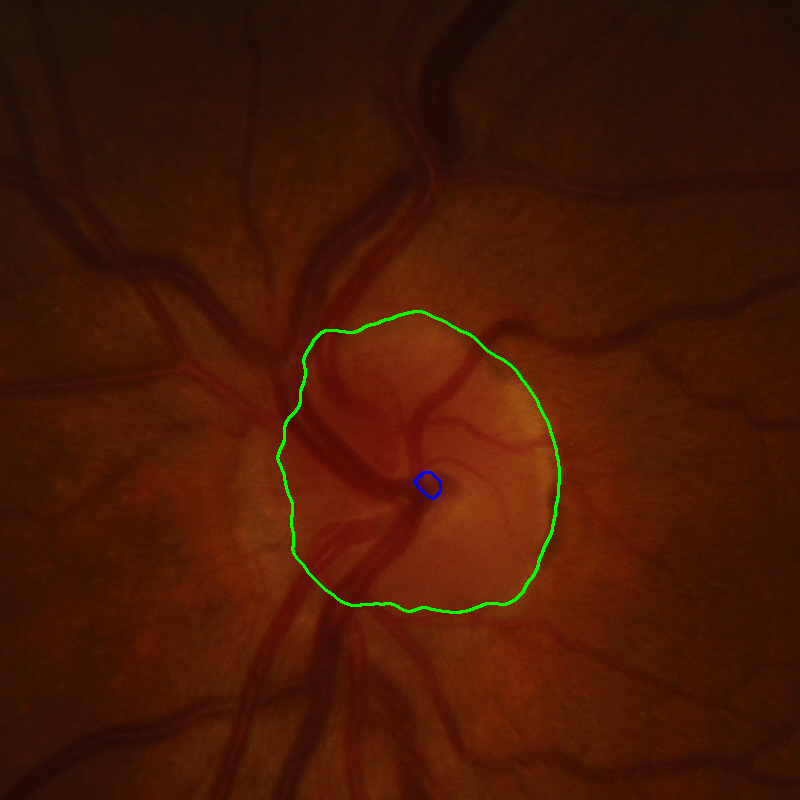}}}&
\bmvaHangBox{\fbox{\includegraphics[width=0.15\linewidth]{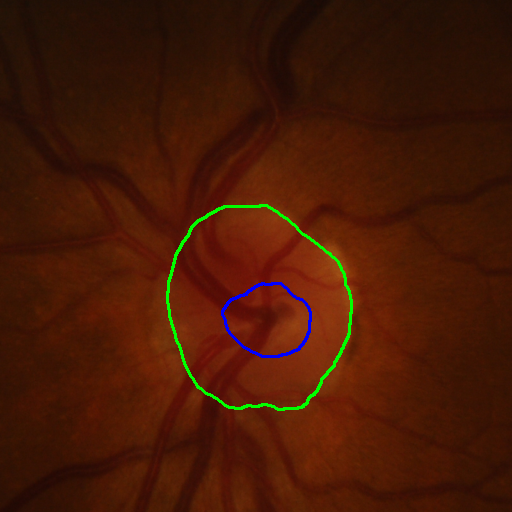}}}\\
(a)~Ground-truth & 
(b)~DPL~\cite{dpl} & 
(c)~CPR~\cite{cpr} & 
(d)~BEAL~\cite{beal} & 
(e)~Grad-CL
\end{tabular}
\vspace{1 em}
\caption{Overlaying the final segmentation on an image from the RIM-ONE-r3 dataset.  
The blue outline denotes the cup and the green outline denotes the disc.  
Competing methods show visibly stretched or contracted regions relative to \textbf{Grad-CL}.}
\label{figureri}
\end{figure*}

\begin{figure*}[t]  
\centering
\begin{tabular}{ccccc}
\bmvaHangBox{\fbox{\includegraphics[width=0.15\linewidth]{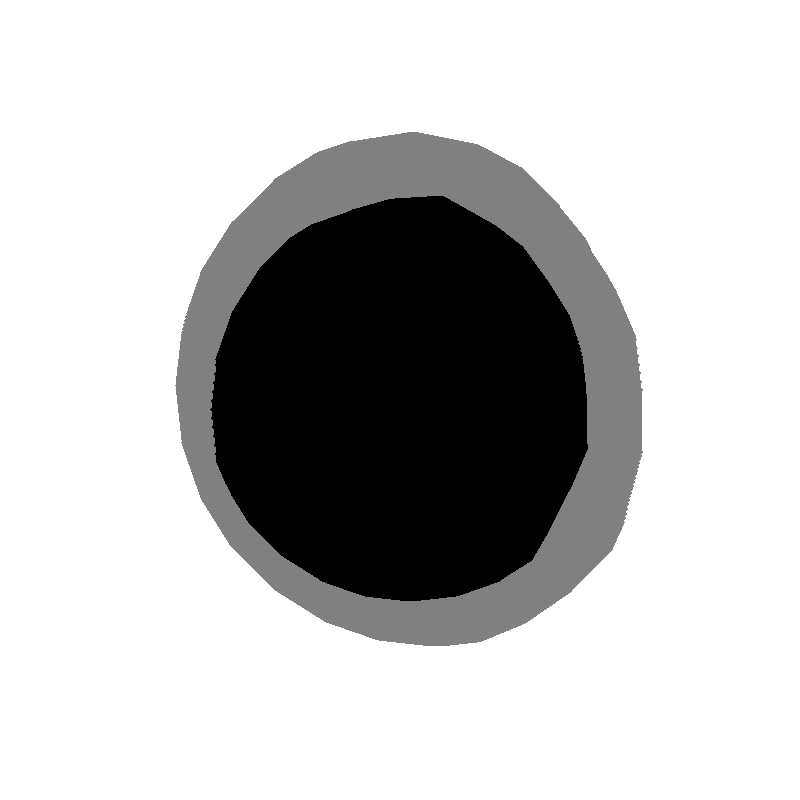}}}&
\bmvaHangBox{\fbox{\includegraphics[width=0.15\linewidth]{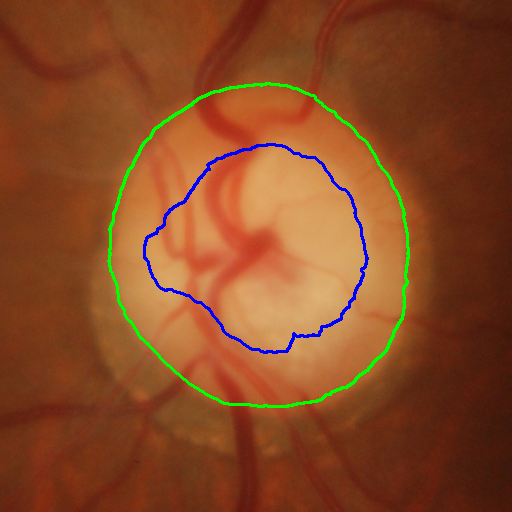}}}&
\bmvaHangBox{\fbox{\includegraphics[width=0.15\linewidth]{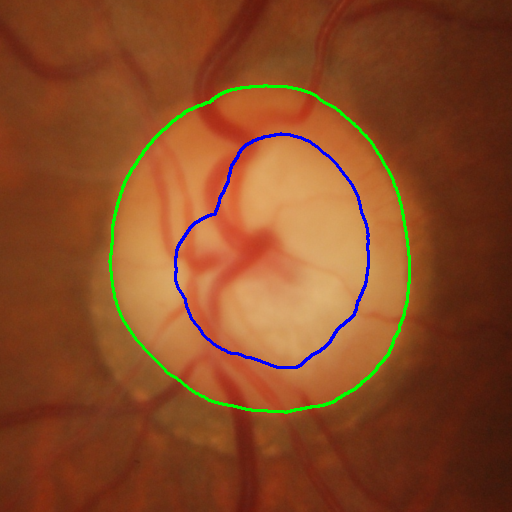}}}&
\bmvaHangBox{\fbox{\includegraphics[width=0.15\linewidth]{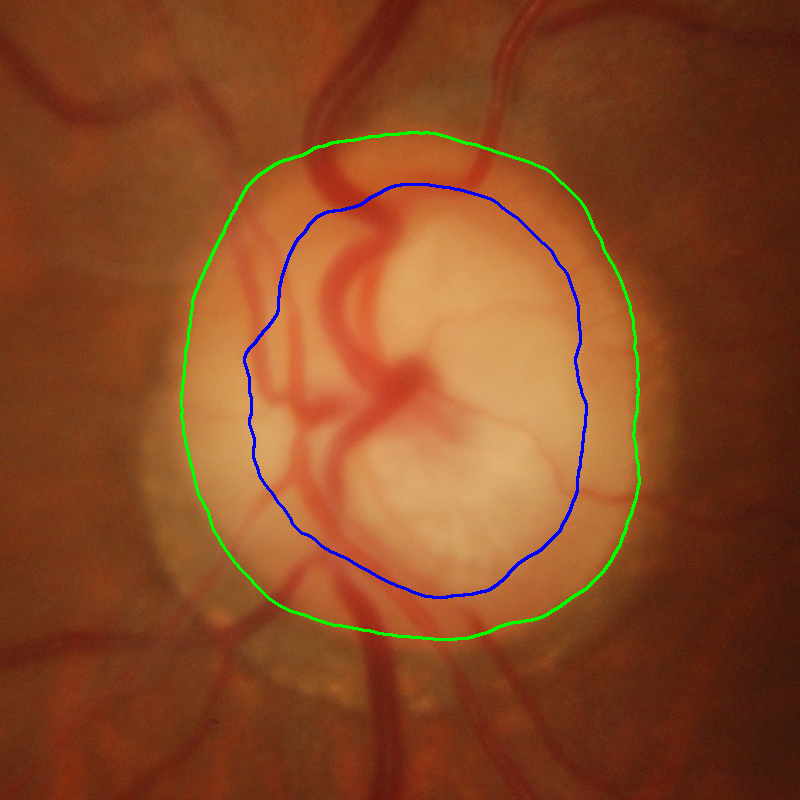}}}&
\bmvaHangBox{\fbox{\includegraphics[width=0.15\linewidth]{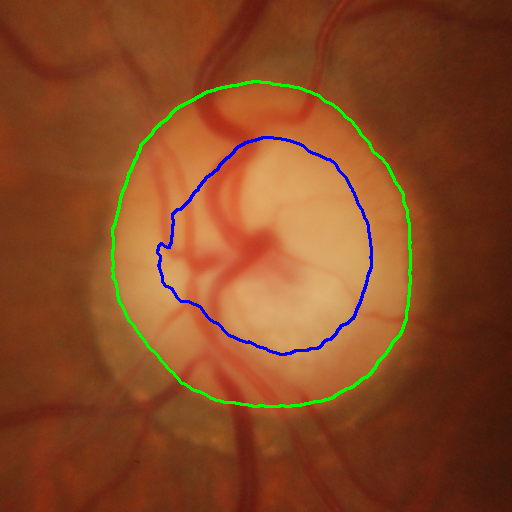}}}\\
(a)~Ground-truth &
(b)~DPL~\cite{dpl} &
(c)~CPR~\cite{cpr} &
(d)~BEAL~\cite{beal} &
(e)~Grad-CL
\end{tabular}
\vspace{1 em}
\caption{Final segmentation overlay on a Drishti-GS image.  
The blue outline marks the optic cup; the green outline marks the optic disc.  
Alternative methods show deformation relative to the proposed \textbf{Grad-CL}.}
\label{figuredrishti}
\end{figure*}

\noindent \textbf{Implementation Details:} The pre-trained source model \(f_s\) generates pseudolabels on target images using a threshold of 0.75 \cite{beal,dpl}. Uncertainty-based refinement employs Monte-Carlo dropout (rate 0.5) with an uncertainty threshold of 0.05 over 10 stochastic passes. For prototype estimation, gradient-highlighted features are used. Our segmentation network is DeepLabv3+ \cite{deeplab} with a MobileNetv2 backbone, trained with the Adam optimizer (momentum 0.9--0.99, learning rate \(2 \times 10^{-3}\)). During adaptation, weak augmentations (random erasing, contrast modification, and Gaussian noise) are applied to slightly perturb inputs and encourage divergence from the pseudolabels. The target model \(f^{s\rightarrow t}\) is trained using \(\mathcal{L}_{total}\) for 20 epochs with a batch size of 8 on PyTorch 1.12.1 and an NVIDIA A100 GPU.
\begin{table}[ht]
\footnotesize
\centering
\begin{tabular}{|ccccc|cccc|}
\hline
\multicolumn{5}{|c|}{\textbf{Drishti-GS}} & \multicolumn{4}{c|}{\textbf{RIM-ONE-r3}} \\ \hline
\multicolumn{1}{|c|}{} & \multicolumn{1}{c|}{Cup} & \multicolumn{1}{c|}{Disc} & \multicolumn{1}{c|}{Cup} & Disc & \multicolumn{1}{c|}{Cup} & \multicolumn{1}{c|}{Disc} & \multicolumn{1}{c|}{Cup} & Disc \\ \hline
\multicolumn{1}{|c|}{Metric} & \multicolumn{2}{c|}{Dice Score} & \multicolumn{2}{c|}{ASD Score} & \multicolumn{2}{c|}{Dice Score} & \multicolumn{2}{c|}{ASD Score} \\ \hline
\multicolumn{1}{|c|}{KL divergence} & \multicolumn{1}{c|}{84.53} & \multicolumn{1}{c|}{96.58} & \multicolumn{1}{c|}{10.38} & 4.05 & \multicolumn{1}{c|}{80.46} & \multicolumn{1}{c|}{95.07} & \multicolumn{1}{c|}{9.58} & 4.98 \\ \hline
\multicolumn{1}{|c|}{JS divergence} & \multicolumn{1}{c|}{84.46} & \multicolumn{1}{c|}{96.53} & \multicolumn{1}{c|}{10.45} & 4.15 & \multicolumn{1}{c|}{79.91} & \multicolumn{1}{c|}{95.24} & \multicolumn{1}{c|}{9.89} & 4.78 \\ \hline
\multicolumn{1}{|c|}{MMD} & \multicolumn{1}{c|}{83.36} & \multicolumn{1}{c|}{96.54} & \multicolumn{1}{c|}{11.23} & 4.13 & \multicolumn{1}{c|}{77.44} & \multicolumn{1}{c|}{94.29} & \multicolumn{1}{c|}{10.49} & 6.11 \\ \hline
\multicolumn{1}{|c|}{Euclidean} & \multicolumn{1}{c|}{80.11} & \multicolumn{1}{c|}{91.51} & \multicolumn{1}{c|}{13.54} & 11.93 & \multicolumn{1}{c|}{63.87} & \multicolumn{1}{c|}{72.63} & \multicolumn{1}{c|}{13.62} & 53.63 \\ \hline
\end{tabular}
\vspace{0.7em}
\caption{ASD and Dice Coefficient on Drishti-GS and RIM-ONE-r3 using different distance/divergence metrics between gradient-based feature maps of the cup and disc.}
\label{tab:my-table2}
\end{table}
\noindent \textbf{Comparative Analysis:} \noindent In this comparative analysis, we perform quantitative evaluations using several baselines and state-of-the-art methods—namely, a without adaptation baseline (lower bound), a supervised adaptation baseline (Oracle) in the target domain, unsupervised domain adaptation (UDA) methods (e.g., BEAL~\cite{beal}, AdvEnt~\cite{advent}, OCDA~\cite{OCDA}, SRDA~\cite{SRDA}, DAE~\cite{DAE}) and source-free domain adaptation (SFDA) methods (e.g., TT-SFUDA~\cite{ttsfda}, TENT~\cite{tent}, DPL~\cite{dpl}, PLPB~\cite{li2023robust}) as reported in Table~\ref{tab:my-table1}; (i) among the UDA methods, BEAL stands out as a benchmark that employs adversarial learning based on boundary predictions and entropy maps between the source and target, making it the best reported UDA method for fundus image segmentation, while AdvEnt is built on entropy minimization, OCDA utilizes a curriculum domain adaptation strategy with memory modules to enhance generalization in open and compound domains, SRDA leverages entropy minimization with a source-trained task prior, and DAE incorporates an implicit prior through denoising autoencoders; (ii) in contrast, most SFDA methods for fundus imaging rely on self-training with pseudo-labeling—with DPL serving as a benchmark for pseudo-label refinement via uncertainty and prototype estimation and PLPB combining entropy minimization with a pseudo-boundary loss for open domain adaptation—yet these approaches tend to fuse the features of the optic cup and disc for prototype estimation, occasionally resulting in misidentification between the two regions; our proposed method, Grad-CL, addresses this limitation by explicitly disaligning cup and disc features through a contrastive loss that utilizes gradient-derived features, thereby enhancing prototype estimation and ultimately outperforming the aforementioned SFDA methods. Most SFDA methods in fundus imaging rely on self-training with pseudolabeling. DPL refines pseudolabels using uncertainty and prototype estimation, while PLPB incorporates entropy minimization with a pseudo-boundary loss. In contrast, Grad-CL not only refines pseudolabels but also disaligns cup and disc features through a contrastive loss, thereby reducing misclassification between these regions.\\
Figures~\ref{figureri} and \ref{figuredrishti} display segmentation overlays for RIM-ONE-r3 and Drishti-GS, respectively, comparing the results of Grad-CL with DPL, CPR, and BEAL.

\noindent \textbf{Ablation Study:}  Table ~\ref{tab:my-table2} evaluates several metrics applied to gradient-based feature maps of the optic cup and disc on the Drishti-GS and RIM-ONE-r3 datasets by comparing divergence metrics (KL and JS divergences) with distance metrics (MMD and Euclidean) using the Dice Coefficient and Average Surface Distance (ASD) as evaluation measures; (i) the divergence metrics yielded high Dice scores (e.g., for Drishti-GS, cup: 84.53 with KL and 84.46 with JS, disc: 96.58 with KL and 96.53 with JS; similar trends were observed on RIM-ONE-r3) along with low ASD values (approximately 10.38–10.45 for the cup and 4.05–4.15 for the disc), and (ii) in contrast, the distance metrics, particularly Euclidean distance, resulted in significantly lower Dice scores and higher ASD values (e.g., on Drishti-GS, cup: 80.11 and disc: 91.51 with Euclidean, accompanied by ASD values up to 13.54 and 11.93, respectively), indicating that enforcing divergence between cup and disc features is more effective than minimizing Euclidean distances in high-dimensional feature spaces; overall, these findings demonstrate that explicitly modeling the divergence between feature distributions not only enhances segmentation performance but can also slightly surpass the performance of traditional cosine similarity losses, especially for optic disc segmentation.

\section{Conclusion}
Grad-CL, a novel SFDA approach for fundus imaging leverages gradient-derived features for enhanced pseudolabel refinement and a contrastive loss to disalign cup and disc features. The inclusion of feature gradient information in prototype estimation and cosine-similarity contrastive loss improves the results. Our method demonstrates performance that is on par with or superior to state-of-the-art source-dependent and source-free approaches. An ablation study validates that enforcing divergence between the gradient-based activations of the cup and disc is crucial for improved segmentation. In future work, we plan to explore advanced gradient-based feature extraction techniques (e.g., Score-CAM, Grad-CAM++, Relevance-CAM) and incorporate source distribution information to further address pronounced domain shifts.
\\
\\
\textbf{Acknowledgments} Rini acknowledges the financial support provided by the Department of Science and Technology, Government of India, through the DST WISE Post-Doctoral Fellowship Program (Reference No. DST/WISE-PDF/ET-33/2023), which facilitated the successful completion of this work. Rajeev acknowledges the support received from the TCS PhD fellowship (Cycle 18). 

\bibliography{egbib}
\end{document}